\documentclass[11pt]{article}

\usepackage[final]{acl}

\usepackage{times}
\usepackage{latexsym}

\usepackage[T1]{fontenc}
\usepackage[utf8]{inputenc}

\usepackage{microtype}

\usepackage{inconsolata}

\usepackage{graphicx}

\usepackage{stfloats}
\usepackage{placeins}
\usepackage{caption}
\usepackage{cuted}

\usepackage{algorithm}
\usepackage{algorithmic}
\usepackage{booktabs}
\usepackage{colortbl}
\usepackage{xcolor}
\usepackage{multirow}
\usepackage{amsmath}
\usepackage{mathtools}
\usepackage{kotex}
\definecolor{slotrow}{HTML}{F4DCC7}  % pale peach for the last row
\usepackage{pifont}% http://ctan.org/pkg/pifont
\usepackage{soul} %\st
\usepackage{amsfonts}
\usepackage{subcaption}
\usepackage{wrapfig}
\definecolor{maroon}{cmyk}{0,0.87,0.68,0.32}
\title{Intrinsic Temporal Adaptation of CLIP for \\ Partially Relevant Video Retrieval}

\author{Hyun Seok Seong \;\;\; Woojin Jun \;\;\; SuBeen Lee \;\;\; Jae-Pil Heo\thanks{Corresponding Author}\\
  Sungkyunkwan University\\
  \texttt{\{gustjrdl95, junwoojin, leesb7426, jaepilheo\}@skku.edu}
  }

\begin{document}
\maketitle

\begin{abstract}
Partially Relevant Video Retrieval~(PRVR) aims to retrieve untrimmed videos that contain moments relevant to a text query.
Since the target moment occupies only a portion of the video, PRVR requires retrieval based on fine-grained understanding beyond coarse video-level matching.
However, existing methods often rely on frozen CLIP frame features, which lack temporal understanding.
Even with recent progress in parameter-efficient CLIP adaptation, video-level predictions can still be supported by imprecise frame-level evidence.
In this paper, we propose an Intrinsic Temporal Adaptation~(ITA) framework for PRVR.
First, our Backbone-Internal Temporal Adaptation allows the last few visual transformer layers to attend over groups of neighboring frames. This provides temporally aware frame embeddings while keeping CLIP frozen and training only adaptation parameters.
Second, we introduce Affinity-Weighted Gradient Propagation to address the weakly supervised nature of PRVR, softly aggregating top-$k$ frames based on text-frame affinities and propagating learning signals to multiple query-relevant frames. 
Our method achieves state-of-the-art performance on PRVR benchmarks, 
demonstrates robust cross-dataset transfer, and retrieves substantially more accurate frame-level evidence within ground-truth query-relevant moments.
Our code is available at \href{github.com/hynnsk/ITA}{github.com/hynnsk/ITA}.

% Recent CLIP-based methods commonly encode frames independently and add temporal modules on top to construct moment-level representations. 
% However, the visual backbone itself remains largely unaware of video-specific temporal dynamics.
% In this paper, we propose a simple yet effective framework that adapts CLIP from within the visual backbone.
% First, our Backbone-Internal Temporal Adaptation allows the last few visual transformer layers to attend over groups of neighboring frames.
% This yields temporally aware frame embeddings while keeping the pretrained backbone frozen with LoRA. 
% Second, we introduce Affinity-Weighted Gradient Propagation to handle weakly supervised nature of PRVR, softly aggregating top-$k$ frames based on text-frame affinities and propagating learning signals to multiple query-relevant frames. 
% Our method achieves state-of-the-art performance on in-domain and cross-domain retrieval performance on PRVR benchmarks. 
% with substantially fewer trainable parameters and lower computation.
\end{abstract}
\section{Introduction}

\begin{figure}[t]
    \centering
    \vspace{-4pt}
    \includegraphics[width=0.98\linewidth]{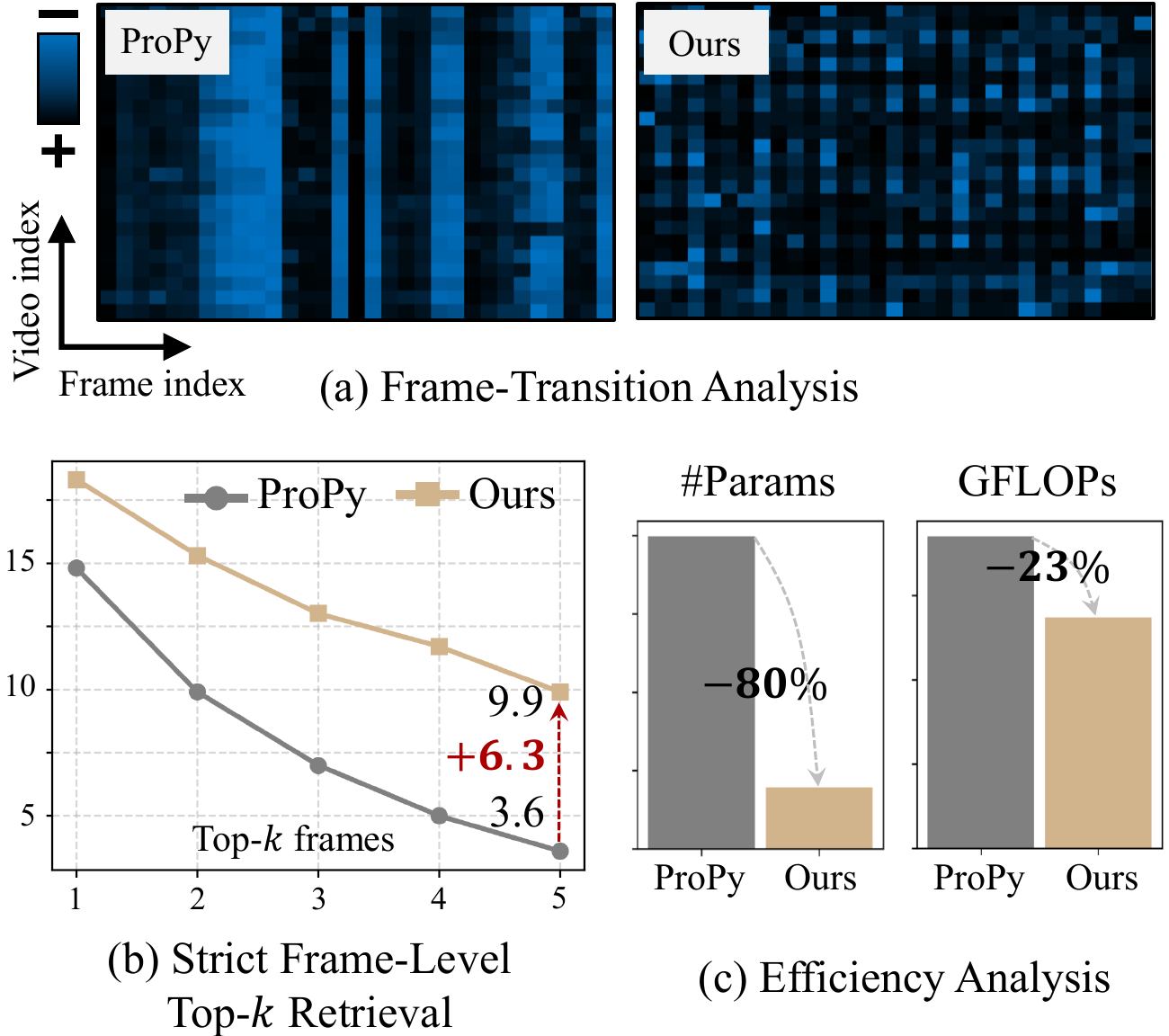}
    \vspace{-4pt}
    % \caption{
    % Analysis comparing ProPy and Ours. (a) For randomly sampled videos, we compute the similarity between the final embedding of each frame and that of its preceding frame.
    % Each row denotes a video and each column denotes a frame index. 
    % (b) Frame-level retrieval performance on ActivityNet Captions under a strict top-$k$ criterion, where a prediction is counted as correct only when all top-$k$ retrieved frames correspond to the query-relevant video.
    % (c) Efficiency comparison in terms of the number of trainable parameters and inference-time GFLOPs of the video encoder.
    % }
    \caption{%
    Comparison between ProPy and Ours.
    (a) Visualization of cosine similarities between each frame and its preceding frame. Rows and columns are video and frame indices, respectively.
    ProPy shows vertical stripe patterns across unrelated videos, while ours produces more video-specific transition patterns.
    (b) Strict frame-level top-$k$ retrieval on ActivityNet Captions.
    Unlike video-level retrieval, we pool frames from all videos and test whether all top-$k$ frames come from the ground-truth video. This evaluates whether video-level retrieval is supported by coherent frame-level evidence. ProPy rapidly degrades as $k$ increases, whereas ours remains more robust.
    (c) Comparison of trainable parameters and inference-time GFLOPs of the video encoder.
    }
    \label{fig:fig_motivation}
\end{figure}

Partially Relevant Video Retrieval~(PRVR) has become increasingly important as large-scale video collections continue to grow~\citep{dong2022partially, jiang2023progressive, dong2023dual, gmmformer, li2025hlformer, msc}.
Unlike conventional text-to-video retrieval, where a text query is assumed to describe the video's global semantics, PRVR considers a more realistic setting in which only a temporal portion of the video is relevant to the query.
% Training PRVR is inherently weakly supervised, as the exact query-related moment is not provided during training.
Therefore, PRVR requires fine-grained understanding of video to identify the evidence that supports text-video matching, rather than relying only on coarse video-level semantics.

To align text queries with relevant visual content, recent PRVR methods commonly adopt CLIP~\citep{clip} as a strong vision-language foundation~\citep{amdnet, propy, msc}.
These methods extract CLIP features from individual video frames and then introduce additional temporal modules to transform frame-level features into moment-level features for text-video matching.
However, these temporal modules must derive moment-level semantics from isolated frame features because CLIP encodes each frame independently without explicit temporal context.
% \HS{기존 방법은 vision embedding을 frame-level과 moment-level 이렇게 two-way로 feature 뽑아서 사용함. 그래서 무거움. }

% \HS{그래서 propy가 효율적으로 하려고 parameter-efficient prompt로 하나의 모델에다가 temporal 정보를 입히려고 시도했다. 하지만 이러이런 문제가 있었음.}
% To reduce the burden of deriving moment-level semantics solely from isolated frame features, ProPy~\citep{propy} augments the CLIP visual encoder with an interactive prompt pyramid, capturing multi-scale temporal information.
% \HS{To reduce the burden of deriving moment-level semantics solely from isolated frame features, ProPy~\citep{propy} augments CLIP with temporal prompt modules.}
To reduce the burden of deriving moment-level semantics solely from isolated frame features, ProPy~\citep{propy} injects learnable temporal prompts into the CLIP visual backbone itself.
This allows temporal cues to be incorporated during feature encoding.
% Rather than using CLIP merely to extract per-frame features and offloading all temporal reasoning to external modules, ProPy~\citep{propy} injects learnable temporal prompts into the CLIP visual backbone itself, so that temporal cues are incorporated during feature encoding rather than only afterward. 이런 느낌? ?? ? ? ?? ? ?? ? ?? 
Despite its progress, we observe that it still does not fully capture video-specific temporal dynamics.
PRVR requires retrieval based on the moment information within a video, but ProPy often fails to provide precise frame-level evidence for its video-level prediction. 
As shown in Fig.~\ref{fig:fig_motivation}~(a), ProPy exhibits position-dependent transition patterns across unrelated videos.
% rather than content-dependent understanding.
This limitation is also reflected in retrieval stability.
As shown in Fig.~\ref{fig:fig_motivation}~(b), when a prediction is counted as correct only if all top-$k$ retrieved frames correspond to the query-relevant video, ProPy's performance drops rapidly as $k$ increases.
This indicates that video-level prediction can still be supported by imprecise frame-level evidence.
% Furthermore, ProPy introduces non-negligible computational overhead, requiring more trainable parameters and higher video-encoder computation~(Fig.~\ref{fig:fig_motivation}~(c)).
% Furthermore, ProPy introduces non-negligible computational overhead, requiring higher video-encoder computation~(Fig.~\ref{fig:fig_motivation}~(c)).

% To address these limitations, we propose an Intrinsic Temporal Adaptation framework for PRVR.
To address these limitations, we propose an Intrinsic Temporal Adaptation~(ITA) framework for PRVR.
First, we introduce Backbone-Internal Temporal Adaptation, which enables the visual encoder to capture video-specific temporal dynamics during feature encoding.
Rather than encoding each frame independently, we allow the last few layers of the encoder to attend over groups of neighboring frames.
As a result, each frame representation is conditioned on its temporal neighbors, incorporating local temporal context into the CLIP visual features.
We then train the adaptation with LoRA~\citep{lora}, while keeping the backbone frozen.

Yet, since PRVR provides only video-level supervision, the text-video score is often computed from the single frame with the highest text-frame similarity~\citep{gmmformerv2, propy, msc}. However, a query-relevant moment may span multiple frames, and the top-1 frame can be noisy or insufficient to represent the whole moment.
To alleviate this limitation, we introduce Affinity-Weighted Gradient Propagation to address the weak supervision in PRVR. Specifically, our method selects and softly aggregates the top-$k$ frames according to text-frame affinities. This allows learning signals to propagate to multiple frames that are likely to belong to the query-relevant moment.

% Second, we introduce Affinity-Weighted Gradient Propagation to address the weak supervision in PRVR. 
% Since PRVR provides only video-level supervision, the text-video score is often computed from the single frame with the highest text-frame similarity~\citep{gmmformerv2, propy, msc}.
% % This concentrates the training signal on that selected frame.
% However, a query-relevant moment may span multiple frames, and the top-1 frame can be noisy or insufficient to represent the whole moment.
% Therefore, our method selects and softly aggregates the top-$k$ frames according to text-frame affinities. This allows learning signals to propagate to multiple frames that are likely to belong to the query-relevant moment.

As shown in Fig.~\ref{fig:fig_motivation}, our design promotes temporally grounded retrieval while maintaining high efficiency.
Experiments on PRVR benchmarks demonstrate that our method achieves strong retrieval performance with a lightweight adaptation of CLIP.

To sum up, our contributions are:
(i) We propose Backbone-Internal Temporal Adaptation, which injects temporal interaction into the CLIP visual encoder without heavy external modules.
(ii) We introduce Affinity-Weighted Gradient Propagation, which propagates contrastive learning signals to multiple high-affinity frames under weak supervision.
% (iii) We demonstrate state-of-the-art PRVR performance with substantially fewer trainable parameters and lower computation.
(iii) We demonstrate state-of-the-art PRVR performance with high efficiency, and further validate backbone generality, cross-dataset transfer, and temporally precise frame-level retrieval.

\section{Related Work}
\label{sec:related_work}

\subsection{\!Text-to-Video Retrieval~(T2VR)}
T2VR aims to identify the video most relevant to a given text query from a large video collection.
Early methods learn cross-modal embeddings to align visual and textual features~\citep{ dong2019dual, liu2019use, chen2020fine, gabeur2020multi}.
With the emergence of CLIP~\citep{clip} as a strong vision-language foundation, recent methods have shifted toward using frame-level CLIP features for video-text matching.
CLIP4Clip~\citep{luo2022clip4clip} establishes a simple and effective paradigm by aggregating frame-level CLIP features into a video representation.
Subsequent works improve CLIP-based T2VR through multi-grained or fine-grained alignment~\citep{ma2022x, liu2022ts2, gorti2022x, wang2022disentangled}, video-language pretraining~\citep{miech2019howto100m, bain2021frozen, xu2021videoclip, wang2022internvideo}, and parameter-efficient adaptation~\citep{xue2022clip, huang2023vop}.
Despite this progress, these methods generally assume global video-text relevance, which often does not hold in a realistic setting where a query may correspond to only a short temporal portion of a video.

\subsection{\!Partially Relevant Video Retrieval~(PRVR)}
PRVR targets a realistic retrieval setting in which a video can be relevant even when only a temporal portion matches the query.
The task is introduced by \citet{dong2022partially}, who employ multi-scale similarity learning to capture clip-level partial relevance.
Subsequent works typically build temporal modules on top of pretrained visual features.
One line of work discovers query-relevant temporal regions through event alignment, Gaussian-based event or clip representations, or learnable span anchors~\citep{jiang2023progressive, gmmformer, gmmformerv2, amdnet}.
Another line improves representation learning under weak supervision through knowledge distillation, prototype-based clip representations, or ambiguity-aware objectives~\citep{dong2023dual, protoprvr, arl}.
Recent works further address semantic collapse~\citep{msc} and hierarchical partial relevance~\citep{li2025hlformer}.
In a different direction, ProPy~\citep{propy} adapts CLIP more explicitly to PRVR by organizing learnable event prompts in a hierarchical structure with cross-granularity interactions.
Despite steady improvements, existing methods commonly treat the visual backbone as a frame-wise feature extractor and rely on top-1 frame aggregation under weak supervision.
In contrast, we enable temporal interaction inside the CLIP backbone and propagate contrastive gradients over multiple high-affinity frames.

\section{Method}

\begin{figure*}[t]
    \centering
    \vspace{-6pt}
    \includegraphics[width=0.98\textwidth]{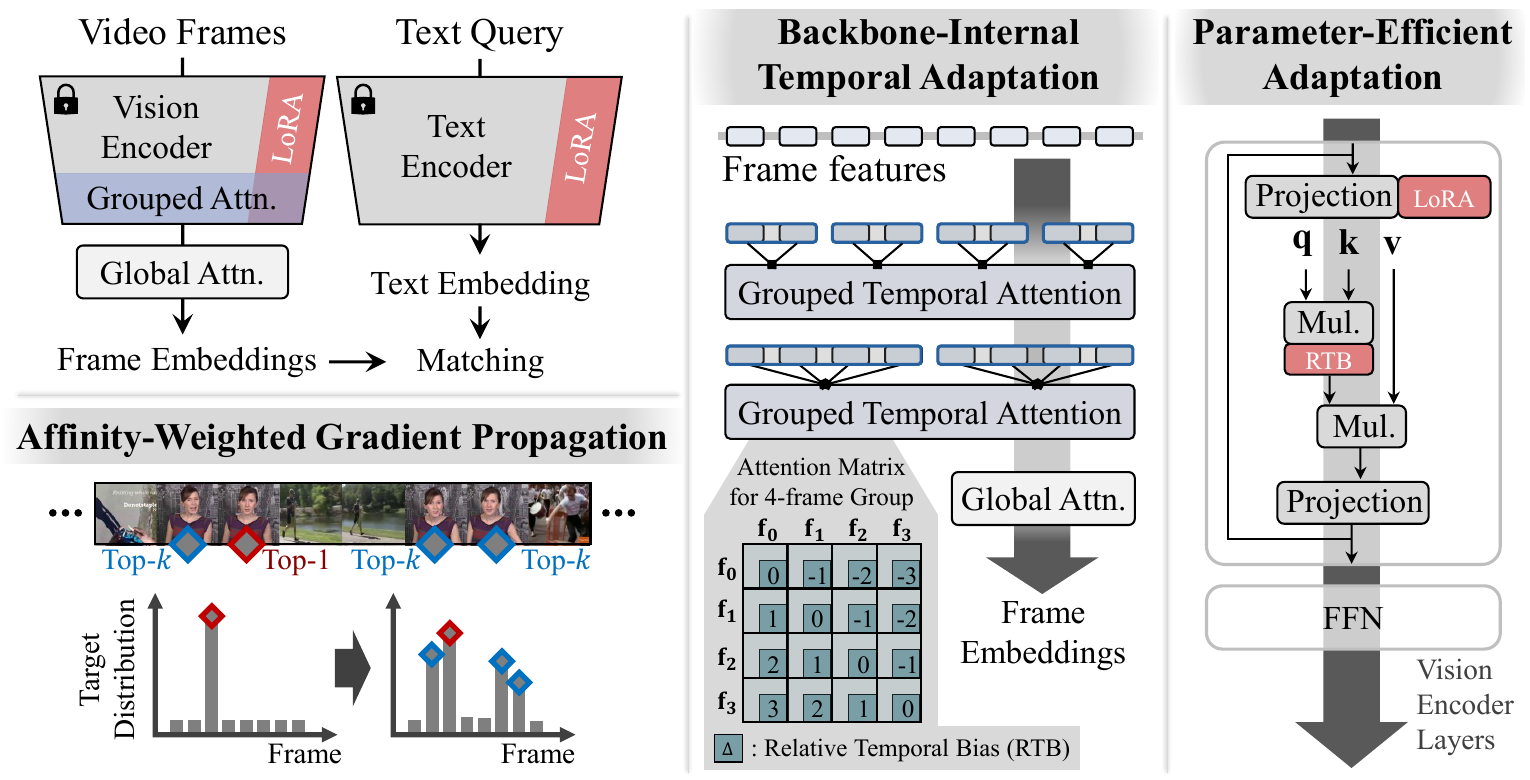}
    \vspace{-8pt}
    \caption{
    Illustration of the proposed framework. Given sampled video frames, the CLIP visual encoder produces frame embeddings through Backbone-Internal Temporal Adaptation, where the last few visual transformer layers perform attention over grouped neighboring frames. The resulting frame embeddings are further refined by a lightweight global class-token attention layer.  To address the weakly supervised nature of PRVR, we introduce Affinity-Weighted Gradient Propagation, which selects top-$k$ frames according to text-frame affinities and propagates contrastive learning signals to multiple query-relevant frames. The whole framework is trained in a parameter-efficient manner by freezing the CLIP backbone and applying LoRA to the query-key-value projection layers.
    }
    \label{fig:fig_main}
    \vspace{-4pt}
\end{figure*}

%%%%%%%%%%%%%%%%%%%%%%%%%%%%%%%%%%%%%%%%%%%%%%%%%%
\subsection{Preliminary}
% method 1을 하는 이유
% method 1에 대한 간략한 설명
% method 2를 하는 이유(PRVR의 weakly-labeled problem)
% method 2에 대한 간략한 설명
% Inference 방식
Given a batch of videos and text queries, our goal is to learn a scoring function that retrieves the video containing a query-relevant moment. 
% Let $\mathcal{B}_v=\{V_j\}_{j=1}^{N_v}$ denote a batch of videos and $\mathcal{B}_q=\{q_i\}_{i=1}^{N_t}$ denote a batch of text queries. 
Each video $V_j$ is uniformly sampled into $T$ frames.
% , i.e., $V_j=\{I_{j,t}\}_{t=1}^{T}$. 
Following the PRVR setting, supervision is provided only at the text-video level: for each query, we are given its positive video index, but the frame or temporal segment relevant to the query is not annotated.

%%%%%%%%%%%%%%%%%%%%%%%%%%%%
\paragraph{Text Representation.}
% CLIP으로 text에 대한 feature 뽑고, <EOS> token을 뽑음
% text-to-video retrieval은 이 <EOS> token으로 비디오를 검색하는것
% Video Encoder와 마찬가지로 LoRA를 attention layer의 in_proj에만 달아서 학습함.
For a text query, we use the CLIP text encoder to obtain its representation. 
The query is tokenized with the standard start and end tokens, and the embedding corresponding to the \texttt{<EOS>} token is projected into the joint embedding space. We denote the query embedding as $\mathbf{t}_i$.

%%%%%%%%%%%%%%%%%%%%%%%%%%%%
\paragraph{Frame Representation.}
For a video $V_j$, the visual encoder produces a sequence of frame-level embeddings $\mathbf{F}_j = [\mathbf{f}_{j,1}, \ldots, \mathbf{f}_{j,T}]$,
% \begin{equation}
%     \mathbf{F}_j
%     =
%     [\mathbf{f}_{j,1}, \ldots, \mathbf{f}_{j,T}]
%     \in \mathbb{R}^{T \times d},
% \end{equation}
where $\mathbf{f}_{j,t}$ denotes the final representation of the $t$-th frame. 
In standard CLIP-based video retrieval, each frame is encoded independently by the CLIP visual encoder, and temporal aggregation is performed outside the backbone. 
In contrast, our method modifies the visual encoding process itself so that $\mathbf{f}_{j,t}$ is temporally contextualized by neighboring frames. 

\paragraph{Contrastive Learning Objective.}
% For training, we first select a frame embedding for each video that has the maximum similarity with query embedding.
Following conventional PRVR training, a baseline text-video score is obtained by selecting the frame embedding with the maximum similarity to the query.
Let $s_{i,j}$ denote the maximum similarity score between text query $\mathbf{t}_i$ and frame embeddings $\mathbf{F}_j$, then we train the model with a bidirectional InfoNCE loss~\citep{simclr, supcon}:
\begin{equation}
    \mathcal{L} = \text{InfoNCE} (s_{i,j})
    \label{eq:infonce}
\end{equation}

% The text-to-video loss is
% \begin{equation}
%     \mathcal{L}_{\mathrm{t2v}}
%     =
%     -\frac{1}{N_t}
%     \sum_{i=1}^{N_t}
%     \log
%     \frac{
%         \exp(s_{i,y_i})
%     }{
%         \sum_{j=1}^{N_v}
%         \exp(s_{i,j})
%     }.
%     \label{eq:t2v_loss}
% \end{equation}
% Since a video may be paired with multiple text queries in the same batch, we define the positive query set of video $V_j$ as
% \begin{equation}
%     \mathcal{P}_{j}
%     =
%     \{i \mid y_i=j\}.
% \end{equation}
% The video-to-text loss is then defined as
% \begin{equation}
%     \mathcal{L}_{\mathrm{v2t}}
%     =
%     -\frac{1}{N_v}
%     \sum_{j=1}^{N_v}
%     \log
%     \frac{
%         \sum_{i \in \mathcal{P}_{j}}
%         \exp(s_{i,j})
%     }{
%         \sum_{i=1}^{N_t}
%         \exp(s_{i,j})
%     }.
%     \label{eq:v2t_loss}
% \end{equation}
% The final objective is
% \begin{equation}
%     \mathcal{L}
%     =
%     \frac{1}{2}
%     \left(
%         \mathcal{L}_{\mathrm{t2v}}
%         +
%         \mathcal{L}_{\mathrm{v2t}}
%     \right).
%     \label{eq:nce_loss}
% \end{equation}
The remaining question is how to compute the frame embeddings $\mathbf{F}_j$ and the text-video score $s_{i,j}$. 
We answer these questions with Backbone-Internal Temporal Adaptation and Affinity-Weighted Gradient Propagation in Sec.~\ref{sec_BITA} and~\ref{sec_AWGP}, respectively.
\subsection{Backbone-Internal Temporal Adaptation} % method 1 이름
\label{sec_BITA}

CLIP is trained on image-text pairs and its visual encoder processes each image independently. 
For PRVR, the representation of a frame should depend on its temporal context, because the relevance of a frame can be determined by neighboring actions or events. 
To facilitate temporal understanding while maintaining parameter efficiency, we adapt the CLIP visual backbone so that temporal interaction occurs inside the backbone.
Unlike external temporal modules that operate on already-extracted frame embeddings, our adaptation enables visual tokens from neighboring frames to interact before the final frame representations are formed.
Specifically, while the original CLIP visual transformer applies self-attention independently to each frame, our method allows the last few layers to attend over tokens from multiple neighboring frames.
% In the original CLIP visual transformer, self-attention is applied independently to the tokens of each frame. 
% In contrast, our method allows the last few layers to attend over tokens from multiple neighboring frames.

%%%%%%%%%%%%%%%%%%%%%%%%%%%%%%%%%%%%%%%%%%
\paragraph{Grouped Temporal Attention.}
For each visual encoder layer $\ell$, we define a frame group size $g_\ell$.
Let $\mathbf{F}^{\ell}_{j,t} \in \mathbb{R}^{L \times d}$ denote the visual tokens of the $t$-th frame in video $V_j$ at layer $\ell$, including the class token and patch tokens.
We group every $g_\ell$ consecutive frames, resulting in the concatenated token sequence $\mathbf{G}^{\ell}_{j,m}\in \mathbb{R}^{g_\ell L \times d}$ for the $m$-th frame group, as follows:
\begin{equation}
    \mathbf{G}^{\ell}_{j,m}
    =
    \mathrm{Cat}
    \left(
    \mathbf{F}^{\ell}_{j,(m-1)g_\ell+1},
    \cdots,
    \mathbf{F}^{\ell}_{j,mg_\ell}
    \right),
\end{equation}
where $m \in \{1,\cdots,M\}$ and $M=\lfloor T/g_\ell \rfloor$.
Then, each grouped token sequence is processed by the standard CLIP self-attention and FFN operations:
\begin{equation}
    \begin{split}
        \!\bar{\mathbf{G}}^{\ell}_{j,m}
        =
        \mathrm{FFN}^{\ell}
        \big(
            \mathrm{MHA}^{\ell}
            (
                \mathbf{G}^{\ell}_{j,m},
                \mathbf{G}^{\ell}_{j,m},
                \mathbf{G}^{\ell}_{j,m}
            )
        \big),
    \end{split}
\end{equation}
where $\mathrm{MHA}^{\ell}$ denotes the multi-head self-attention in the $\ell$-th encoder layer, with the grouped token sequence as query, key, and value.
The output is then split back into frame-wise token sequences:
\begin{equation}
    \!\!
    \left[
    \mathbf{F}^{\ell+1}_{j,(m-1)g_\ell+1},
    \cdots,
    \mathbf{F}^{\ell+1}_{j,mg_\ell}
    \right]
    \!\!=\!
    \mathrm{Split}_{g_\ell}
    (
    \bar{\mathbf{G}}^{\ell}_{j,m}
    ),
\end{equation}
where $\mathrm{Split}_{g_\ell}$ divides the grouped token sequence into $g_\ell$ frame-wise sequences.

In practice, we apply grouped temporal attention only to the last few visual layers~(e.g., $g_\ell \in \{2,4\}$ for the last two layers).
This is motivated by the observation that early layers mainly capture local visual patterns, while deeper layers encode higher-level semantics~\citep{interpreting, multimodal} that are more suitable for temporal event understanding.

Although grouped temporal attention enables local temporal interaction, each group still has a finite temporal receptive field. 
To further incorporate video-level context across all sampled frames, we apply a lightweight global frame attention over the final class-token embeddings.  
Since this refinement operates only on frame-level class-token embeddings rather than dense patch tokens, it adds only a small computational cost.
Overall, this design gives the frame-level output features temporal awareness while preserving the simplicity and efficiency of the CLIP backbone. 
The resulting frame embeddings $\mathbf{F}_j=[\mathbf{f}_{j,1},\ldots,\mathbf{f}_{j,T}]$ are used for retrieval.

%%%%%%%%%%%%%%%%%%%%%%%%%%%%%%%%%%%%%%%%%%
\paragraph{Temporal Bias.}
% Positional Embedding with Relative Temporal Bias
To make grouped temporal attention aware of frame order, we add two lightweight temporal signals. 
First, we use a learnable temporal positional embedding $\mathbf{e}_t \in \mathbb{R}^{d}$ for each frame index $t$, which is added to all tokens from the same frame.
Second, for layers with $g_\ell>1$, we apply a learnable relative temporal bias, extending position bias in Swin Transformer~\citep{swin} to temporal dimension.
For two tokens $u$ and $v$ inside the same grouped sequence, let $\gamma(u)$ and $\gamma(v)$ denote their frame indices within the group. 
The attention logit is modified as:
\begin{equation}
    a_{u,v}
    =
    \frac{
        \mathbf{q}_{u}
        \mathbf{k}_{v}
        ^{\top}
    }{
        \sqrt{d}
    }
    +
    r_{\gamma(u)-\gamma(v)},
    \label{eq:relative_temporal_bias}
\end{equation}
where $d$ is the feature dimension and $r_{\Delta}$ is a learnable scalar for relative temporal offset $\Delta \in \{-(g_\ell-1),\ldots,g_\ell-1\}$. 
This bias enables the attention layer to distinguish tokens not only by visual content but also by their relative temporal positions.

%%%%%%%%%%%%%%%%%%%%%%%%%%%%%%%%%%%%%%%%%%
\paragraph{Parameter-Efficient Adaptation.}
% LoRA로 fine tuning한다는 설명
We freeze the pretrained CLIP backbone and train only lightweight adaptation parameters.
Specifically, we apply LoRA~\citep{lora} to the query-key-value input projection layer of every attention block in both the visual and text encoders.
% For every attention block in both the visual and text encoders, we apply LoRA to the packed query-key-value projection matrix. 
Because the original input projection, output projection, MLP, and normalization parameters remain frozen, the adaptation is highly parameter-efficient. 
In addition to LoRA, we train only the temporal positional embeddings, relative temporal biases, and the lightweight global frame attention.

%%%%%%%%%%%%%%%%%%%%%%%%%%%%%%%%%%%%%%%%%%%%%%%%%%
\subsection{Affinity-Weighted Gradient Propagation} % method 2 이름
\label{sec_AWGP}
% Relevance-Aware
% Affinity-Weighted
PRVR provides only video-level supervision, although the text query is usually relevant to a temporal subset of the video. 
A common approach for PRVR computes the text-video similarity by selecting the maximum text-frame similarity.
% \begin{equation}
%     s^{\mathrm{top1}}_{i,j}
%     =
%     \max_{t}
%     \mathbf{t}_i^{\top}\mathbf{f}_{j,t}.
% \end{equation}
While simple, this top-1 aggregation sends the contrastive learning signal primarily to a single selected frame. 
This is suboptimal for PRVR because a query-relevant moment often spans multiple frames, and the top-1 frame can be noisy or insufficient to represent the whole event.

We therefore introduce an objective tailored to the weakly labeled nature of PRVR. 
Specifically, we select the top-$k$ frames according to text-frame affinity, softly aggregate them, and apply a bidirectional NCE loss to the resulting video representation. 
Formally, for each query-video pair, we first compute frame-level affinities:
\begin{equation}
    a_{i,j,t}
    =
    \mathbf{t}_i^{\top}
    \mathbf{f}_{j,t}.
\end{equation}
We then select the corresponding top-$k$ frames:
\begin{equation}
    \mathcal{K}_{i,j}
    =
    \mathrm{TopK}_{t}
    \left(
    a_{i,j,t}
    \right).
\end{equation}
% In implementation, the top-$k$ indices are selected using stop-gradient affinities to avoid unstable optimization through the discrete selection operation.

For the selected frames, we compute affinity weights:
\begin{equation}
    w_{i,j,t}
    =
    \frac{
    \exp\left(a_{i,j,t}/\tau\right)
    }{
    \sum_{u \in \mathcal{K}_{i,j}}
    \exp\left(a_{i,j,u}/\tau\right)
    },
    \quad
    t \in \mathcal{K}_{i,j},
\end{equation}
where $\tau$ is a temperature parameter.
The final video embedding is then obtained by a weighted sum of the selected frame embeddings:
% \begin{equation}
%     {\mathbf{h}}_{i,j}
%     =
%     \mathrm{Normalize}
%     \left(
%     \sum_{t \in \mathcal{K}_{i,j}}
%     w_{i,j,t}
%     \mathbf{f}_{j,t}
%     \right).
% \end{equation}
\begin{equation}
    {\mathbf{h}}_{i,j}
    =
    \mathrm{Normalize}
    \left(
    \sum\nolimits_{t \in \mathcal{K}_{i,j}}
    w_{i,j,t}
    \mathbf{f}_{j,t}
    \right).
\end{equation}
Finally, the text-video score is computed as
\begin{equation}
    s_{i,j}
    =
    \mathbf{t}_i^{\top}
    {\mathbf{h}}_{i,j}.
    \label{eq:awgp_score}
\end{equation}
% where $\gamma$ is the CLIP logit scale.

When $k>1$, the NCE gradient is distributed to multiple high-affinity frames. 
Thus, the model is encouraged to align the query not only with the single most discriminative frame but also with other frames that plausibly belong to the same query-relevant moment.

We use the score in Eq.~\ref{eq:awgp_score} in the bidirectional NCE objective defined in Eq.~\ref{eq:infonce}.
% No auxiliary supervision or moment-level annotation is required. 
The final training objective remains a standard contrastive loss, while its gradient is propagated in a relevance-aware manner over multiple candidate frames.

Importantly, although $k$ is fixed, the effective supervision of AWGP remains query-adaptive. The selected set $\mathcal{K}_{i,j}$ is determined independently for each query-video pair according to their relative affinities, and the softmax weights continuously modulate the contribution of each selected frame. Consequently, relatively weak candidates can receive negligible weights even when they are included in the top-$k$ set. 
Furthermore, each selected frame embedding has already been contextualized through Grouped Temporal Attention and global frame attention, such that gradients applied to a selected embedding can also propagate through its temporal interactions. Thus, using a fixed $k$ does not restrict the effective learning signal to an equally weighted set of $k$ isolated frames.

%%%%%%%%%%%%%%%%%%%%%%%%%%%%%%%%%%%%%%%%%%%%%%%%%%
\subsection{Inference}

During inference, we use the same scoring function as in training.
For each query-video pair, we compute frame-level affinities, select the top-$k$ frames, aggregate them with affinity weights, and obtain the final score using Eq.~\ref{eq:awgp_score}.
Videos are then ranked according to this score.
Compared with the conventional top-1 aggregation, the additional cost of top-$k$ selection and affinity-weighted aggregation is negligible.
\section{Experiment}

\begin{table*}[t]
\centering
\caption{Performance comparison on TVR, ActivityNet Captions, and Charades-STA. Rows highlighted in gray indicate the performance of methods leveraging ResNet152 + I3D + RoBERTa features.}
\vspace{-8pt}
\label{tab:performance_comparison}

\renewcommand{\arraystretch}{1.0} % Default value: 1
\setlength{\tabcolsep}{3.5pt} % Default value: 6pt

\resizebox{\textwidth}{!}{%
\begin{tabular}{l|ccccc|ccccc|ccccc}
\toprule
\multicolumn{1}{c|}{Method}
& \multicolumn{5}{c|}{TVR}
& \multicolumn{5}{c|}{ActivityNet Captions}
& \multicolumn{5}{c}{Charades-STA} \\
& R@1 & R@5 & R@10 & R@100 & SumR
& R@1 & R@5 & R@10 & R@100 & SumR
& R@1 & R@5 & R@10 & R@100 & SumR \\
\midrule

\rowcolor{gray!20}
MS-SL~\citep{dong2022partially} & 13.5 & 32.1 & 43.4 & 83.4 & 172.4
& 7.1 & 22.5 & 34.7 & 75.8 & 140.1
& 1.8 & 7.1 & 11.8 & 47.7 & 68.4 \\

\rowcolor{gray!20}
MS-SL++~\citep{chen2025prvr} & 13.6 & 33.1 & 44.2 & 83.5 & 174.5
& 7.0 & 23.1 & 35.2 & 75.8 & 141.1
& 1.8 & 7.6 & 12.0 & 48.4 & 69.7 \\

\rowcolor{gray!20}
PEAN~\citep{jiang2023progressive} & 13.5 & 32.8 & 44.1 & 83.9 & 174.2
& 7.4 & 23.0 & 35.5 & 75.9 & 141.8
& 2.7 & 8.1 & 13.5 & 50.3 & 74.7 \\

\rowcolor{gray!20}
GMMFormer~\citep{gmmformer} & 13.9 & 33.3 & 44.5 & 84.9 & 176.6
& 8.3 & 24.9 & 36.7 & 76.1 & 146.0
& 2.1 & 7.8 & 12.5 & 50.6 & 72.9 \\

\rowcolor{gray!20}
DL-DKD~\citep{dldkd} & 14.4 & 34.9 & 45.8 & 84.9 & 179.9
& 8.0 & 25.0 & 37.5 & 77.1 & 147.6
& - & - & - & - & - \\

\rowcolor{gray!20}
Proto~\citep{protoprvr} & 15.4 & 35.9 & 47.5 & 86.4 & 185.1
& 7.9 & 24.9 & 37.2 & 77.3 & 147.4
& - & - & - & - & - \\

\rowcolor{gray!20}
ARL~\citep{arl} & 15.6 & 36.3 & 47.7 & 86.3 & 185.9
& 8.3 & 24.6 & 37.4 & 78.0 & 148.3
& - & - & - & - & - \\

\rowcolor{gray!20}
GMMFormer-V2~\citep{gmmformerv2} & 16.2 & 37.6 & 48.8 & 86.4 & 189.1
& 8.9 & 27.1 & 40.2 & 78.7 & 154.9
& 2.5 & 8.6 & 13.9 & 53.2 & 78.2 \\

\midrule

MS-SL~\citep{dong2022partially} & 20.1 & 43.8 & 55.0 & 89.4 & 208.3
& 11.9 & 32.6 & 45.3 & 82.3 & 172.1
& 1.5 & 5.9 & 9.6 & 43.3 & 60.3 \\

MS-SL++~\citep{chen2025prvr} & 23.4 & 47.2 & 58.3 & 90.1 & 219.0 
& 12.7 & 33.1 & 46.0 & 82.1 & 173.9 
& 1.5 & 5.2 & 9.3 & 39.2 & 55.2 \\

GMMFormer~\citep{gmmformer} & 20.5 & 44.1 & 55.8 & 90.2 & 210.5
& 13.5 & 33.7 & 47.0 & 82.1 & 176.3 
& 1.1 & 5.6 & 9.7 & 43.6 & 60.0 \\

GMMFormer-V2~\citep{gmmformerv2} & 22.6 & 46.5 & 58.5 & 91.6 & 219.1
& 13.3 & 33.9 & 47.4 & 82.8 & 177.4
& 1.2 & 5.5 & 9.5 & 42.9 & 59.1 \\

AMDNet~\citep{amdnet} & 21.0 & 44.2 & 56.2 & 90.0 & 211.4 
& 11.5 & 32.0 & 44.9 & 82.0 & 170.5 
& 1.5 & 5.8 & 10.2 & 42.8 & 60.3 \\

HLFormer~\citep{li2025hlformer} & 22.6 & 46.7 & 58.0 & 91.2 & 218.4 
& 13.1 & 34.2 & 47.3 & 82.9 & 177.4 
& 1.2 & 4.9 & 9.1 & 42.4 & 57.6 \\

MSC~\citep{msc} & 22.3 & 46.3 & 57.5 & 91.0 & 217.2
& 12.5 & 33.1 & 45.9 & 82.5 & 174.0
& 1.9 & 7.2 & 11.3 & 47.1 & 67.4 \\

ProPy~\citep{propy} & 22.4 & 45.0 & 55.9 & 89.5 & 212.8
& 14.9 & 34.9 & 47.5 & 82.7 & 180.0
& 2.9 & 9.8 & 15.0 & 54.8 & 82.5 \\

\rowcolor{maroon!10}
ITA~(Ours) & 25.3 & 49.2 & 59.8 & 90.5 & 224.7 & 19.2 & 41.7 & 54.4 & 86.1 & 201.4 & 3.5 & 10.6 & 16.8 & 56.5 & 87.4 \\

\bottomrule

\end{tabular}%
}
\end{table*}

%%%%%%%%%%%%%%%%%%%%%%%%%%%%%%%%%%%%%%%%%%%%%%%%%%%%
\subsection{Experiment Settings}
\paragraph{Datasets and Metrics.}
We evaluate our method on four widely used PRVR benchmarks: TVR~\citep{tvr}, ActivityNet Captions~\citep{activitynet}, Charades-STA~\citep{charades}, and QVHighlights~\citep{qvhighlights}. 
% These datasets contain untrimmed or temporally diverse videos paired with natural language queries, making them suitable for evaluating whether a retrieval model can identify videos that contain query-relevant moments.
% Each video is associated with multiple queries, with an average of 3.3 text queries per video, where different queries often refer to semantically distinct moments. 
TVR contains 17,435 videos and 87,175 queries for training, and 2,179 videos and 10,895 queries for evaluation. 
For ActivityNet Captions, each video is paired with an average of 3.7 text queries, with 10,009 videos used for training and 4,917 videos used for evaluation. 
Charades-STA contains 13,898 video-sentence pairs for training and 4,233 pairs for evaluation.
QVHighlights consists of videos collected from news and vlog-style content.
% , and has been reorganized for the PRVR setting in prior work~\citep{protoprvr}.

For evaluation, we use recall-based retrieval metrics. 
Specifically, we report R@$K$, which measures the percentage of text queries whose ground-truth video is ranked within the top-$K$ retrieved results. We also report SumR, the sum of all recall scores.

\paragraph{Implementation Details.}
We use CLIP-B/32~\cite{clip} as the backbone for both the visual and text encoders. 
For each video, we uniformly sample $T=32$ frames. 
The temporal receptive fields of grouped temporal attention are set to $\{2,4\}$ in the final visual transformer layers.
For parameter-efficient adaptation, we set the LoRA~\citep{lora} rank to 8.
For Affinity-Weighted Gradient Propagation, $k$ and $\tau$ are set to 4 and 0.05, respectively.
These hyperparameter settings are used across all datasets.
We train all models for 10 epochs with a batch size of 48. 
The learning rate is set to $2\times10^{-4}$ for ActivityNet Captions and Charades-STA, $1\times10^{-3}$ for TVR, and $1\times10^{-4}$ for QVHighlights. 
All experiments are conducted on a single NVIDIA RTX A6000 GPU with an Intel Xeon Gold 5220R~(2.20GHz) CPU.
The training time is approximately  2.0, 1.5, 0.5, and 1.5 hours for TVR, ActivityNet Captions, Charades-STA, and QVHighlights, respectively.

%%%%%%%%%%%%%%%%%%%%%%%%%%%%%%%%%%%%%%%%%%%%%%%%%%%%
\subsection{Comparison with the State of the Art}
% CLIP-B/32 실험

\begin{table}[t]
\small
\centering
\caption{Performance on QVHighlights \textit{val} split. Rows highlighted in gray are results with CLIP-B/16 features.}
\vspace{-8pt}
\label{tab:qvhighlights_val}
\renewcommand{\arraystretch}{1.0} % Default value: 1
\setlength{\tabcolsep}{3.5pt} % Default value: 6pt
\begin{tabular}{l|ccccc}
\toprule
Method & R@1 & R@5 & R@10 & R@100 & SumR \\
\midrule
\rowcolor{gray!15}
GMMFormer & 18.2 & 43.7 & 56.7 & 92.5 & 211.1 \\
\rowcolor{gray!15}
MS-SL     & 20.4 & 46.7 & 60.7 & 94.6 & 222.5 \\
\rowcolor{gray!15}
Proto     & 22.6 & 48.8 & 61.3 & 93.9 & 226.6 \\
\midrule
% GMMFormer    & 16.3 & 39.7 & 52.3 & 88.4 & 196.7 \\
% AMDNet       & 17.1 & 40.8 & 52.5 & 88.4 & 198.8 \\
% GMMFormer-V2 & 15.6 & 40.2 & 53.7 & 88.5 & 198.0 \\
% MS-SL        & 17.4 & 43.4 & 55.2 & 88.8 & 204.8 \\
% ProPy & 37.4 & 65.6 & 76.1 & 96.5 & 275.5 \\
% \midrule
MS-SL        & 21.2 & 50.0 & 61.0 & 94.1 & 226.4 \\
GMMFormer    & 18.4 & 42.3 & 56.2 & 93.4 & 210.3 \\
GMMFormer-V2 & 23.1 & 50.7 & 63.9 & 94.8 & 232.5 \\
AMDNet       & 21.4 & 47.2 & 59.8 & 93.2 & 221.5 \\
HLFormer     & 21.0 & 45.9 & 59.9 & 93.7 & 220.5 \\
MSC          & 20.7 & 46.8 & 59.5 & 93.7 & 220.7 \\
ProPy        & 37.4 & 65.6 & 76.1 & 96.5 & 275.5 \\
\rowcolor{maroon!10}
ITA~(Ours) & 39.8 & 67.2 & 78.0 & 96.8 & 281.9 \\
\bottomrule
\end{tabular}
\end{table}

% 4개의 데이터셋에 대해서 베이스라인들과 비교함.
% 베이스라인들을 CLIP-B/32 백본 기준, 32 프레임 샘플링으로 reproduce했다.
% 우리 방법이 제일 좋은 성능을 달성했다. 
% 특히 가장 보편적인 scene이고 real-world 데이터셋인 activitynet-captions에서 높은 성능을 달성했다. SumR 기준 ProPy보다 21.4 상승했음. TVR에서의 성능 향상도 상당하고, Charades와 QV에서도 비교적 modest하긴 하지만 state-of-the-art 달성했음
% 우리의 temporal understanding behavior가 더 정밀한 visual understanding 을 만들었고 이게 좋은 성능을 만듦.
We compare our method with existing PRVR baselines on four benchmarks. 
For a fair comparison under the same backbone setting, we reproduce the baselines using CLIP ViT-B/32 with 32 sampled frames per video~(except MS-SL++~\cite{chen2025prvr}). 
As shown in Tab.~\ref{tab:performance_comparison} and \ref{tab:qvhighlights_val}, our method achieves the best overall performance.
In particular, our method shows a large improvement on ActivityNet Captions, a diverse and real-world video dataset, improving SumR over ProPy~\citep{propy} by 21.4 points. 
The gain on TVR is also substantial, demonstrating the effectiveness of our method in retrieving partially relevant moments from long and semantically diverse videos. 
Our method also achieves state-of-the-art performance on Charades-STA and QVHighlights.
These results suggest that our temporally grounded frame representations enable more precise visual understanding, leading to stronger PRVR performance.
% \paragraph{Generalization across Backbones.}
Furthermore, to examine whether the proposed adaptation is specific to CLIP ViT-B/32, we further evaluate it with a CLIP variant and video-pretrained video-language backbones. As shown in Tab.~\ref{tab:backbone}, our method consistently improves over ProPy across CLIP ViT-B/16, CLIP4Clip-B/32~\cite{luo2022clip4clip}, and InternVideo-MM-B/16~\cite{wang2022internvideo}. In particular, the improvement with InternVideo indicates that our adaptation remains complementary even when the pretrained representation already incorporates temporal information.

%%%%%%%%%%%%%%%%%%%%%%%%%%%%%%%%%%%%%%%%%%%%%%%%%%%%
\subsection{Ablation Study}
\paragraph{Component Ablation.}
% Ablation Study (component)

We conduct an ablation study to evaluate the contribution of each component in our framework.
As shown in Tab.~\ref{tab:component_ablation}, the baseline with only Parameter-Efficient Adaptation provides a competitive starting point.
Adding Grouped Temporal Attention consistently improves performance, showing that temporal interaction inside the CLIP visual backbone is beneficial for PRVR.
Temporal Bias further improves the model, suggesting that temporal order and relative offsets help grouped attention better model frame representations.
Affinity-Weighted Gradient Propagation also improves performance when added to the PEA baseline, indicating that propagating supervision to multiple high-affinity frames is effective under weak video-level supervision.
Combining all components yields the best performance, demonstrating that temporal representation learning and affinity-weighted training are complementary.

\paragraph{Hyperparameter Ablation.}
% Ablation Study (k 같은것들)

% \begin{table}[t]
% \centering
% \small
% \setlength{\tabcolsep}{6pt}
% \caption{Generalization across visual/video-language backbones. 
% We report the average SumR over four PRVR benchmarks.}
% \label{tab:backbone}
% \begin{tabular}{lcc}
% \toprule
% Backbone & ProPy & Ours \\
% \midrule
% CLIP ViT-B/16       & 198.0 & \textbf{209.0} \\
% CLIP4Clip-B/32      & 183.6 & \textbf{196.1} \\
% InternVideo-MM-B/16 & 190.7 & \textbf{201.7} \\
% \bottomrule
% \end{tabular}
% \end{table}

\begin{table}[t]
\centering
\small
\setlength{\tabcolsep}{6pt}
\caption{Generalization across diverse backbones. 
We report the average SumR over four PRVR benchmarks.}
\vspace{-6pt}
\label{tab:backbone}
\begin{tabular}{lcc}
\toprule
Backbone & ProPy & \cellcolor{maroon!10} ITA \\
\midrule
CLIP-B/16       & 198.0 & \cellcolor{maroon!10} 209.0 \\
CLIP4Clip-B/32      & 183.6 & \cellcolor{maroon!10} 196.1 \\
InternVideo-MM-B/16 & 190.7 & \cellcolor{maroon!10} 201.7 \\
\bottomrule
\end{tabular}
\vspace{-2pt}
\end{table}

\begin{table}[t]
\centering
\caption{Ablation study on each component, reporting SumR. PEA, GTA, TB, and AW denote Parameter-Efficient Adaptation, Grouped Temporal Attention, Temporal Bias, and Affinity-Weighting, respectively.}
\vspace{-8pt}
\label{tab:component_ablation}
\renewcommand{\arraystretch}{1.0} % Default value: 1
\setlength{\tabcolsep}{4pt} % Default value: 6pt
\resizebox{\linewidth}{!}{
\begin{tabular}{cccc|cccc|c}
\toprule
PEA & GTA & TB & AW & TVR & ANet & Charades & QV & Avg.\\
\midrule
\checkmark &            &            &           & 217.7   & 195.9   & 78.5 & 271.2 & 190.8 \\
\checkmark & \checkmark &            &           & 221.1 & 199.0  & 82.0 & 277.6    & 194.9 \\
\checkmark & \checkmark & \checkmark &            & 223.2  & 199.6& 83.9 & 277.9    & 196.2 \\
\checkmark &            &            & \checkmark & 219.6  & 199.5      & 81.7   & 280.8    & 195.4 \\
\checkmark & \checkmark & \checkmark & \checkmark & 224.7 & 201.4 & 87.4 & 281.9 & 198.9 \\
\bottomrule
\end{tabular}
}
\end{table}

\begin{figure}[t]
    \centering
    \includegraphics[width=0.98\linewidth]{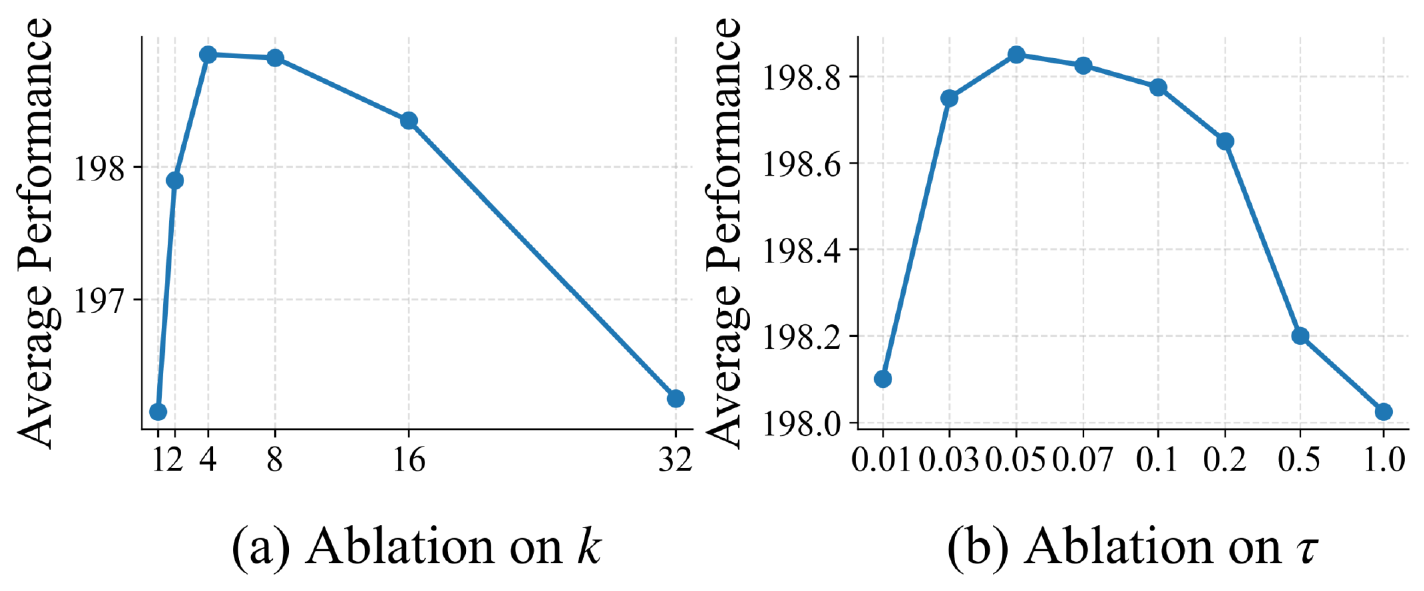}
    \vspace{-6pt}
    \caption{
    Hyperparameter ablation on the number of selected frames $k$ and the affinity temperature $\tau$. 
    For $k$, we fix $\tau=0.05$ and vary $k$ from top-1 to all-frame aggregation.
    For $\tau$, we fix $k=4$ and vary $\tau$ from 0.01 to 1.0. 
    We report average SumR over four datasets.
    }
    \label{fig:fig_ablation_hyperparameter}
\end{figure}

We study the effect of $k$ and $\tau$ in Affinity-Weighted Gradient Propagation.
As shown in Fig.~\ref{fig:fig_ablation_hyperparameter}, selecting multiple high-affinity frames outperforms conventional top-1 aggregation.
However, using too many frames gradually degrades performance, as irrelevant frames can be included and dilute the relevance-aware supervision.
The temperature $\tau$ controls the sharpness of affinity weights among the selected frames.
Since top-$k$ candidates often have similar affinity scores, a proper temperature is needed to emphasize more reliable frames.
A very small $\tau$ makes the aggregation close to top-1 selection, whereas a very large $\tau$ makes the weights nearly uniform and may over-emphasize less relevant frames.
Based on this, we use $k$=4 and $\tau$=0.05 as the default setting.

\paragraph{Temporal Receptive Field Ablation.}
% 레이어별 프레임 몇개 합칠건지 vision encoder의 GFLOPS랑 같이
% \input{table/table_00_temporal_receptive_field_ablation}

\begin{figure}[t]
    \centering
    \includegraphics[width=0.85\linewidth]{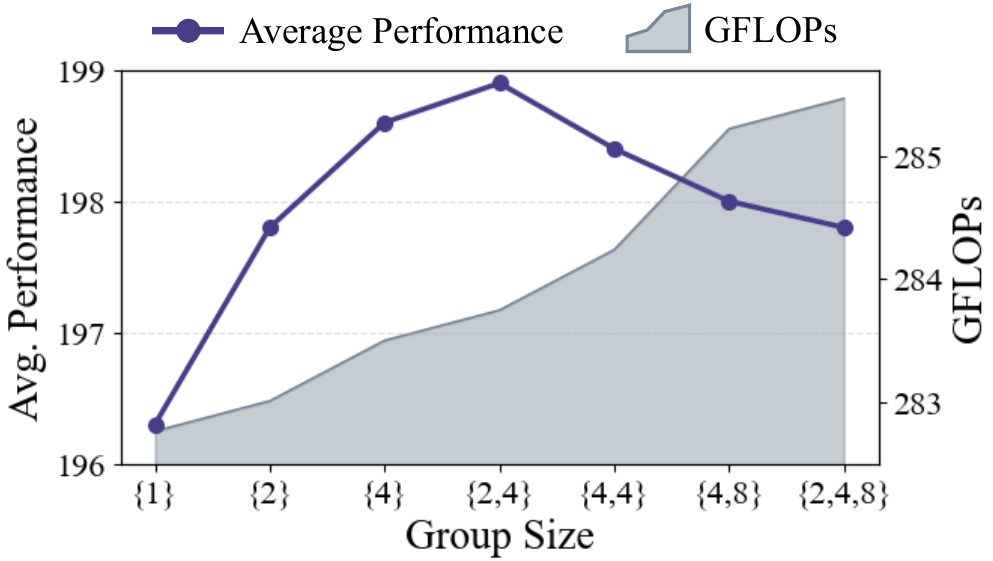}
    \vspace{-6pt}
    \caption{
    Ablation study on the group size for grouped temporal attention.
    GFLOPs denotes the video encoding cost. We report the average SumR over the four datasets.
    }
    \label{fig:ablation_group_size}
\end{figure}

\begin{figure}[t]
    \centering
    \includegraphics[width=0.98\linewidth]{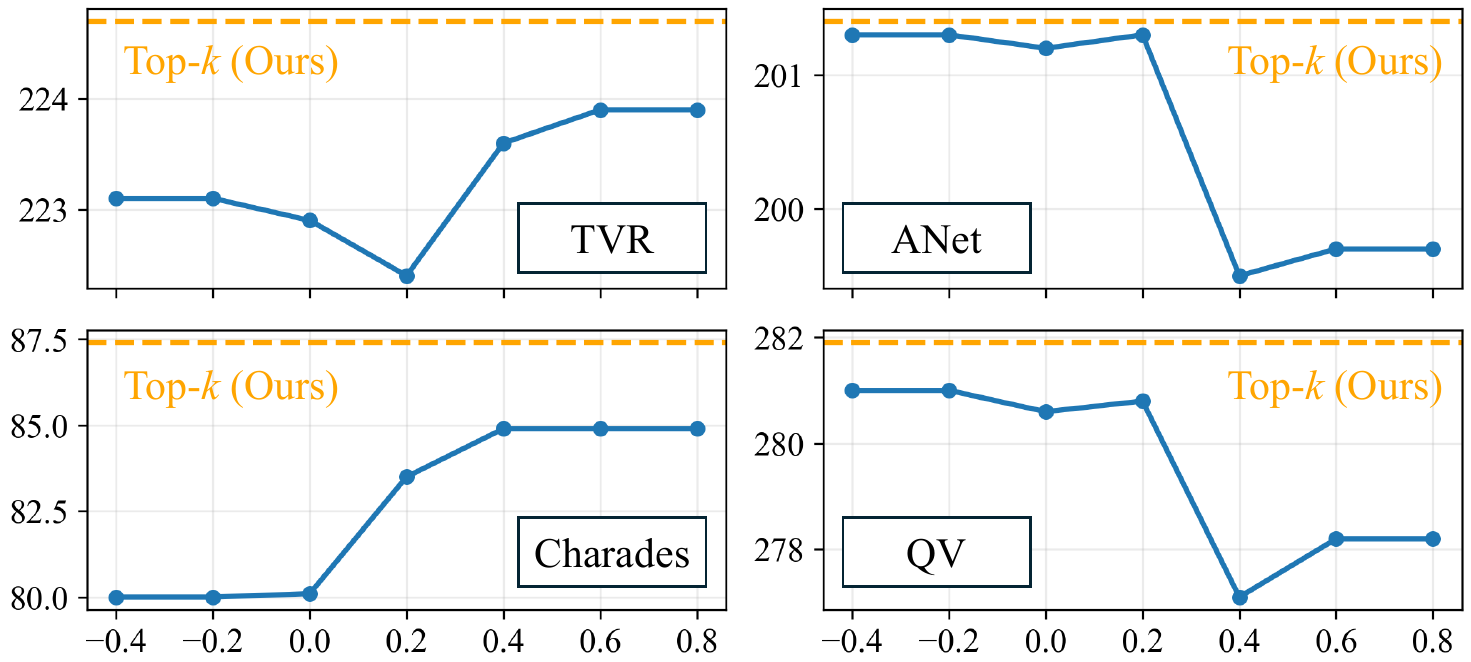}
    \vspace{-6pt}
    \caption{
    Comparison between top-$k$ and threshold-based frame selection.
    The orange dashed lines indicate our final model, while the blue curves show threshold-based selection using different fixed affinity 
    thresholds~(x-axis). The y-axis denotes SumR.
    }
    \vspace{-6pt}
    \label{fig:fig_threshold_based}
\end{figure}

We also study different temporal grouping configurations in grouped temporal attention.
As shown in Fig.~\ref{fig:ablation_group_size}, temporal grouping improves performance over the frame-independent setting, but the gain does not monotonically increase with larger groups or more adapted layers.
% This suggests that simply expanding the temporal receptive field is not always beneficial, since overly broad grouping may mix weakly related frames and weaken frame-level specificity.
This suggests that overly broad temporal interaction may weaken frame-level specificity.
% Therefore, effective temporal adaptation requires a balance between incorporating temporal context and preserving precise frame-level evidence.
We therefore choose $\{2,4\}$ as our default configuration, balancing retrieval performance and computation.
% Based on this trade-off between performance and computation, we choose the $\{2,4\}$ configuration as our default setting.
In addition to the ablation above, we further test sensitivity to predefined group boundaries using overlapping \{4, 4\} and sliding \{2, 4\} grouping. Overlapping grouping obtains an average SumR of 198.2, indicating limited sensitivity to the exact boundary placement. Sliding grouping slightly improves the average SumR to 199.5, but increases video-encoding cost from 283.7 to 313.6 GFLOPs because each frame is repeatedly processed in overlapping windows. We therefore retain the non-overlapping configuration for its favorable accuracy-efficiency trade-off.

\paragraph{Top-$k$ vs. Threshold-Based Selection.}
% One potential alternative to fixed top-$K$ selection is affinity-threshold filtering. We therefore replace top-$K$ selection with a fixed 
% affinity threshold and vary the threshold from $-0.4$ to $0.8$.
We compare fixed top-$k$ selection with affinity-threshold filtering by varying the threshold from $-0.4$ to $0.8$.
As shown in Fig.~\ref{fig:fig_threshold_based}, no single threshold consistently outperforms our top-$k$ strategy across the four datasets. This is mainly because absolute text-frame cosine similarities are not equally calibrated across datasets and queries, making a globally fixed threshold sensitive to affinity distribution. In contrast, top-$k$ selection depends on the relative ranking of frame affinities and is therefore more stable across different datasets. Importantly, fixed $k$ does not imply that the selected frames contribute equally: AWGP adaptively weights the selected frames for each query-video pair and suppresses relatively weak candidates. Thus, top-$k$ provides stable selection while retaining query-adaptive frame weighting.

%%%%%%%%%%%%%%%%%%%%%%%%%%%%%%%%%%%%%%%%%%%%%%%%%%%%
\subsection{Cross-dataset generalization}

We evaluate the generalization ability of PRVR models under a cross-dataset setting.
As shown in Tab.~\ref{tab:cross_dataset}, all methods are trained only on QVHighlights and then directly evaluated on other PRVR benchmarks.
% This setting is challenging because the target datasets differ substantially in video domain, visual style, event structure, and language distribution.
Most existing methods show limited transferability to unseen datasets. 
% Although they can learn effective retrieval patterns on the source dataset, their performance drops noticeably when evaluated on different domains.
This suggests that many PRVR models tend to overfit to dataset-specific temporal patterns or annotation biases.
% , rather than learning broadly transferable video-text representations.
Although ProPy shows stronger generalization than earlier methods due to its CLIP-based adaptation, its transfer performance is still limited.
% Our method achieves the best performance across all datasets. 
% This demonstrates that adapting temporal understanding inside the CLIP visual backbone leads to more transferable frame representations. 
Conversely, our approach not only improves in-domain PRVR performance but also generalizes more robustly to unseen video domains.
These results indicate that our method learns task-relevant frame features instead of relying on dataset-specific temporal proposals or heavy interaction modules.

% \begin{table}[t]
% \centering
% \caption{asd}
% \label{tab:temporal_group_ablation}
% \resizebox{\linewidth}{!}{
% \begin{tabular}{l|c|ccc}
% \toprule
% Method & ANet & TVR & Charades & QV  \\
% \midrule
% MS-SL & 172.1 & 53.7 & 35.0 & 199.6 \\
% GMMFormer & 176.3 & 47.3 & 35.1 & 188.2 \\
% GMMFormer-V2 & 177.4 & 49.0 & 34.7 & 191.1 \\
% AMDNet & 170.5 & 45.8 & 36.7 & 194.8 \\
% HLFormer & 177.4 & 49.1 & 35.6 & 192.0 \\
% MSC & 174.0 & 44.4 & 32.4 & 188.6 \\
% ProPy & 180.0 & 102.2 & 50.4 & 250.7 \\
% Ours & 201.4 & 121.8 & 57.4 & 259.2 \\
% \bottomrule
% \end{tabular}
% }
% \end{table}

\begin{table}[t]
\small
\centering
\caption{Cross-dataset generalization results. 
All methods are trained only on QVHighlights and evaluated on the other datasets.
QV denotes the source-domain evaluation and we report SumR for all results.}
\vspace{-8pt}
\label{tab:cross_dataset}
% \resizebox{\linewidth}{!}{
\begin{tabular}{l|c|ccc}
\toprule
Method & QV & TVR & ANet & Charades\\
\midrule
MS-SL & 226.4 & 41.2 & 96.5 & 30.6 \\
GMMFormer & 210.3 & 38.1 & 91.1 & 28.7 \\
% GMMFormer-V2 & 232.5 & 6.4 & 4.9 & 10.3 \\
AMDNet & 221.5 & 37.2 & 86.7 & 30.4 \\
HLFormer & 220.5 & 39.2 & 92.2 & 29.6 \\
MSC & 220.7 & 40.5 & 94.2 & 27.8 \\
ProPy & 275.5 & 102.5 & 150.3 & 46.0 \\
\rowcolor{maroon!10}
ITA~(Ours) & 281.9 & 120.3 & 173.5 & 60.9 \\
\bottomrule
\end{tabular}
% }
\end{table}

\begin{figure*}[t]
    \centering
    \vspace{-10pt}
    \includegraphics[width=0.98\linewidth]{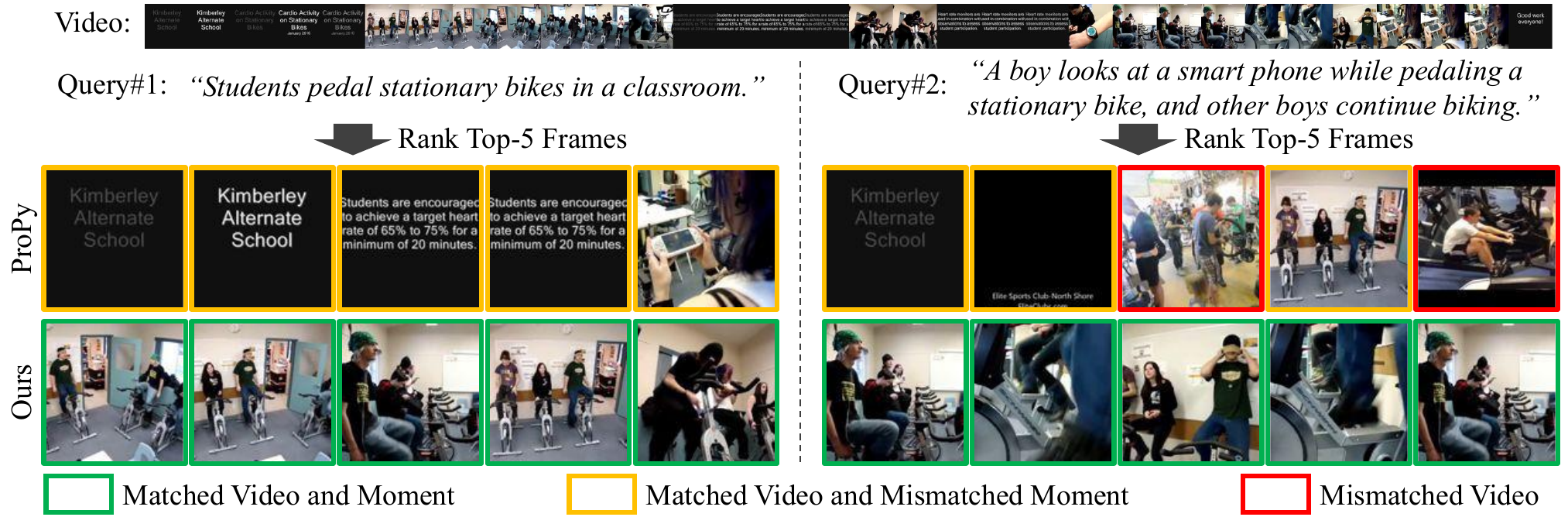}
    \vspace{-6pt}
    \caption{
    Qualitative comparison of frame-level retrieval. 
    Given a text query, we first perform video-level retrieval and consider examples where both ProPy and our method retrieve the correct video. 
    We then rank individual frames and visualize the top-5 retrieved frames for each method.
    }
    \vspace{-6pt}
    \label{fig:fig_frame_level_retrieval}
\end{figure*}

%%%%%%%%%%%%%%%%%%%%%%%%%%%%%%%%%%%%%%%%%%%%%%%%%%%%
% \subsection{Qualitative Results}
% Qualitative results

%%%%%%%%%%%%%%%%%%%%%%%%%%%%%%%%%%%%%%%%%%%%%%%%%%%%
% \subsection{Generalization across Backbones}
% CLIP-L/14 실험
% 근데 문제가 CLIP-L lora든 adapter든 finetuning할때 8GPU 듦 ㅋ..

%%%%%%%%%%%%%%%%%%%%%%%%%%%%%%%%%%%%%%%%%%%%%%%%%%%%
\subsection{Analysis on Temporal Understanding}
% 비디오 축으로 뭔갈 더 잘 이해한다는 분석 있으면 좋을듯.
% dyGMM모듈, ProPY랑 비교

% \begin{figure}[t]
%     \centering
%     \includegraphics[width=0.98\linewidth]{figure/fig_transition_heatmap_v3.pdf}
%     % \vspace{-0.2cm}
%     \caption{
%     Visualization of consecutive-frame similarity.
%     Each row corresponds to a randomly sampled video and each column corresponds to a frame index; brighter colors indicate lower similarity of adjacent frames. 
%     }
%     \label{fig:fig_transition_statistics}
% \end{figure}

\begin{figure}[t]
    \centering
    \includegraphics[width=0.98\linewidth]{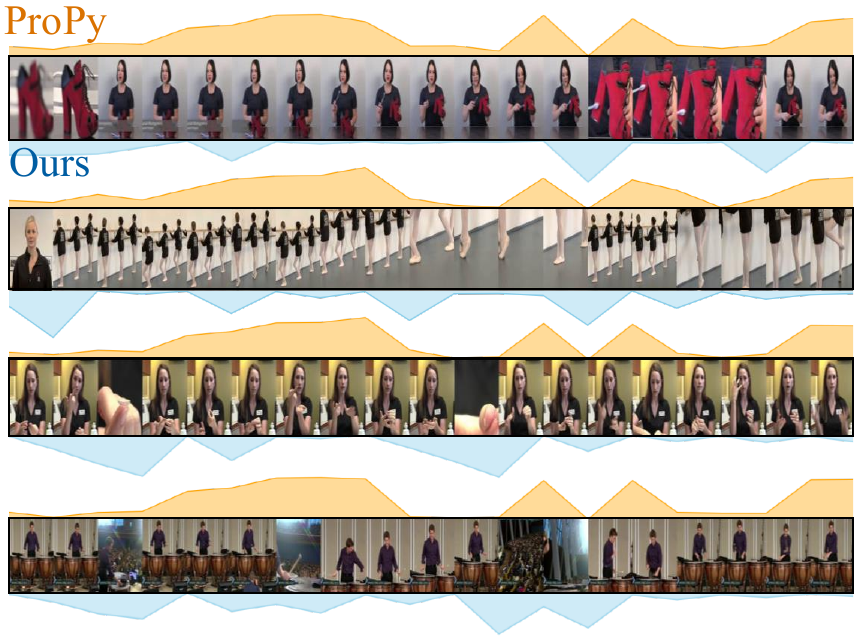}
    \vspace{-8pt}
    \caption{
    Visualization of frame-transition score for frames (4th-22nd frames). Orange and blue curves correspond to ProPy and Ours, respectively.
    }
    \vspace{-8pt}
    \label{fig:fig_transition_graph}
\end{figure}

\paragraph{Frame-Level Retrieval.}
% While standard PRVR evaluation focuses on video-level retrieval, a correct video-level prediction does not necessarily imply that the model has identified the correct temporal evidence inside the video.
% Therefore, we conduct a qualitative frame-level retrieval analysis to examine whether a model retrieves a partially relevant video based on the actual query-relevant moment.
% We further analyze whether correct video-level retrieval is supported by accurate frame-level evidence.
We further examine whether video-level retrieval is supported by accurate frame-level evidence.
Fig.~\ref{fig:fig_motivation}~(b) evaluates top-$k$ frame retrieval on ActivityNet Captions, counting a prediction as correct only when all top-$k$ frames come from the query-relevant video.
% ProPy rapidly degrades as $k$ increases, showing that its video-level prediction is often not supported by precise frame-level matching.
ProPy degrades rapidly as $k$ increases, suggesting that its video-level retrieval is often supported by imprecise frame matches.
We additionally evaluate temporal grounding using ground-truth moment annotations, where a retrieved frame is correct only if it lies within the annotated query-relevant moment.
% To directly measure temporal grounding, we additionally use the ground-truth moment annotations available in the benchmarks. Specifically, we rank all frames in the video database according to their similarity to a query, and count a retrieved frame as correct only when it falls inside the annotated query-relevant moment. 
% Tab.~\ref{tab:frame_retrieval} shows that our method consistently outperforms ProPy across all four datasets and all recall levels.
As shown in Tab.~\ref{tab:frame_retrieval}, our method consistently outperforms ProPy across all datasets and recall levels.
% Fig.~\ref{fig:fig_frame_level_retrieval} also visualizes examples where both methods retrieve the correct video.
% Although ProPy succeeds at the video level, its top-ranked frames often correspond to irrelevant moments or even mismatched videos.
% In contrast, our method retrieves frames that are directly aligned with the text query.
Fig.~\ref{fig:fig_frame_level_retrieval} also shows that, even when both methods retrieve the correct video, our method identifies frames more directly aligned with the query.
This provides qualitative evidence that our method performs PRVR based on temporally grounded frame understanding.

% \begin{table}[t]
% \centering
% \small
% \setlength{\tabcolsep}{6pt}
% \caption{Frame-level retrieval using ground-truth query-relevant moments.}
% \label{tab:frame_retrieval}
% \begin{tabular}{llccc}
% \toprule
% Dataset & Method & R@1 & R@5 & R@10 \\
% \midrule
% ANet
% & ProPy & 5.61 & 13.03 & 18.31 \\
% \rowcolor{maroon!10}
% & Ours  & \textbf{9.86} & \textbf{18.99} & \textbf{24.64} \\
% \midrule
% TVR
% & ProPy & 6.95 & 16.26 & 21.63 \\
% \rowcolor{maroon!10}
% & Ours  & \textbf{11.90} & \textbf{23.95} & \textbf{31.37} \\
% \midrule
% Char.
% & ProPy & 0.67 & 2.80 & 4.14 \\
% \rowcolor{maroon!10}
% & Ours  & \textbf{2.12} & \textbf{4.70} & \textbf{6.75} \\
% \midrule
% QV
% & ProPy & 23.23 & 45.48 & 55.68 \\
% \rowcolor{maroon!10}
% & Ours  & \textbf{32.52} & \textbf{52.52} & \textbf{60.06} \\
% \bottomrule
% \end{tabular}
% \end{table}

\begin{table}[t]
\centering
\small
\setlength{\tabcolsep}{6pt}
\caption{Frame-level retrieval using GT moments.}
\vspace{-6pt}
\label{tab:frame_retrieval}
\begin{tabular}{llccc}
\toprule
Dataset & Method & R@1 & R@5 & R@10 \\
\midrule
TVR
& ProPy & 6.95 & 16.26 & 21.63 \\
& \cellcolor{maroon!10} ITA
& \cellcolor{maroon!10} 11.90
& \cellcolor{maroon!10} 23.95
& \cellcolor{maroon!10} 31.37 \\
\midrule
ANet
& ProPy & 5.61 & 13.03 & 18.31 \\
& \cellcolor{maroon!10} ITA
& \cellcolor{maroon!10} 9.86
& \cellcolor{maroon!10} 18.99
& \cellcolor{maroon!10} 24.64 \\
\midrule
Charades
& ProPy & 0.67 & 2.80 & 4.14 \\
& \cellcolor{maroon!10} ITA
& \cellcolor{maroon!10} 2.12
& \cellcolor{maroon!10} 4.70
& \cellcolor{maroon!10} 6.75 \\
\midrule
QV
& ProPy & 23.23 & 45.48 & 55.68 \\
& \cellcolor{maroon!10} ITA
& \cellcolor{maroon!10} 32.52
& \cellcolor{maroon!10} 52.52
& \cellcolor{maroon!10} 60.06 \\
\bottomrule
\end{tabular}
\vspace{-6pt}
\end{table}

\paragraph{Frame-Transition Analysis.}

To analyze whether the learned frame representations capture video-specific temporal dynamics, we conduct frame-transition analysis.
Given the final frame embeddings, we compute the similarity between each frame and its immediately preceding frame.
Fig.~\ref{fig:fig_motivation}~(a) visualizes the transition patterns over randomly sampled videos, and Fig.~\ref{fig:fig_transition_graph} shows representative examples.
Ideally, transition patterns should vary according to the actual video content.
However, ProPy shows similar transition responses at fixed frame positions across videos, suggesting position-dependent artifacts or collapsed temporal behavior.
In contrast, our method produces more diverse and video-dependent patterns, indicating that it better preserves frame-level temporal variations.
These results suggest that our method learns temporally grounded frame representations and performs retrieval based on meaningful video dynamics.

%%%%%%%%%%%%%%%%%%%%%%%%%%%%%%%%%%%%%%%%%%%%%%%%%%%%
\subsection{Efficiency}

% dyGMM모듈, ProPY랑 비교
% training parameter, inference GFLOPS 등 비교
% MSC 등 기존 워크랑 효율성 비교 (inference GFLOPS 등)
%%% 두가지버전으로. (1)피쳐까지 뽑는속도 (2)피쳐다뽑아놓은가정에서 속도
% 검색 속도랑 메모리 비교

We compare the efficiency of our method with baselines in terms of trainable parameters and computational cost at inference in Tab.~\ref{tab:efficiency}.
All computational costs are measured under the same input setting with
CLIP ViT-B/32 and 32 sampled frames per video.
ProPy~\citep{propy} reduces trainable parameters and online retrieval cost compared with prior methods, but it still requires a relatively high video encoding cost.
This matters in practical retrieval systems, where large-scale video content must be continuously encoded and indexed.
Our method achieves a better efficiency-performance trade-off.
Compared with ProPy, our method uses substantially fewer trainable parameters and reduces the offline video encoding cost by 23.3\%, while achieving the highest average SumR.
It also reduces the online retrieval cost by 10.1\% when searching over 10,000 pre-extracted videos. 
Since our method does not introduce a heavy text-side module or query-dependent video re-encoding, it remains efficient in both offline feature construction and online retrieval.
% These results show that our method improves both offline feature construction and online retrieval efficiency while maintaining strong retrieval performance.

% 근데 our method achieves large reduction of trainable parameters compared to ProPy, owing to its lightweight adaptation strategy without heavy interaction modules.
% ProPy는 근데 video encoding에서 많은 연산을 필요로 한다.
% 이건 important for practical retrieval systems where large-scale video content must be continuously encoded and indexed.
% 우리는 also reduces the offline video encoding cost by 23.3\%.
% % , which is important for practical retrieval systems where large-scale video content must be continuously encoded and indexed.
% The advantage is also observed in the online retrieval setting. When retrieving from a database of 10,000 pre-extracted videos, it reduces GFLOPs by 10.1\%.
% % The online cost is determined by text-query encoding and database matching. 
% Our method does not introduce a heavy text-side module or query-dependent video re-encoding, so its online retrieval cost remains close to the lower bound of a CLIP-based system.
% These results demonstrate that our method improves efficiency in both offline feature construction and online retrieval, while maintaining a simple and lightweight architecture.

\begin{table}[t]
\centering
\small
\caption{
Efficiency comparison with existing PRVR methods. 
GFLOPs (video) denotes the offline cost of encoding a single video.
GFLOPs (online) denotes the online retrieval cost when visual embeddings of 10,000 database videos are pre-extracted and stored.
}
\vspace{-8pt}
\renewcommand{\arraystretch}{1.0} % Default value: 1
\setlength{\tabcolsep}{6pt} % Default value: 6pt
\begin{tabular}{l@{\hspace{-2pt}}ccc|c}
\toprule
Method
& Trainable & GFLOPs & GFLOPs & Avg.\\
& Params~(M) $\downarrow$ & (video) $\downarrow$ & (online) $\downarrow$ & SumR \\
\midrule
MS-SL & 2.70 & 282.8 & 10.285 & 166.8\\
% GMMFormer & 10.72 & 283.4 & 6.236 & 164.3\\
GMM-V2 & 30.11 & 284.7 & 6.475 & 172.0\\
% AMDNet & 0.89 & 282.7 & 6.000 & 165.9 \\
% HLFormer & 27.74 & 284.6 & 6.475 & 168.5\\
MSC & 30.24 & 284.0 & 6.289 & 169.8\\
ProPy & 7.98 & 370.0 & 6.107 & 187.7\\
\rowcolor{maroon!10}
ITA  & 1.57 & 283.7 & 5.490 & 198.9\\
\bottomrule
\end{tabular}
\vspace{-8pt}
\label{tab:efficiency}
\end{table}

\section{Conclusion}
In this paper, we proposed an Intrinsic Temporal Adaptation framework for PRVR. First, our Backbone-Internal Temporal Adaptation injects temporal interaction directly into the CLIP visual encoder. Second, Affinity-Weighted Gradient Propagation distributes contrastive learning signals to multiple high-affinity frames under weak supervision. 
Experiments on four PRVR benchmarks show that our method achieves state-of-the-art performance with fewer trainable parameters and lower video-encoder computation. 
Additional analyses further demonstrate that our method learns more temporally grounded frame representations and generalizes better across datasets.

\section*{Limitations}
% Our method is built on CLIP and inherits its representation biases. 
% These biases can affect retrieval results and can cause vicious cycle where precise frame-level label did not provided because PRVR is inherent weakly-supervised. CLIP의 inherent bias를 넘어선 retrieval system이 추후 과제이지 않을까 싶다.
Our method is built on CLIP and may inherit its representation biases. 
Since PRVR provides only video-level supervision without precise frame-level labels, these biases can be reinforced during training, causing the model to favor frames that are highly aligned with CLIP but not necessarily the most temporally accurate evidence. 
Overcoming such biases under weak supervision remains an important direction for future work.

\section*{Potential Risks}
% The potential risks of this work are similar to those of general video retrieval systems. 
Improved retrieval models can help users search large video collections more efficiently, but they may also be misused for surveillance, privacy-invasive search, or large-scale indexing of sensitive video content. 
Responsible deployment should consider data privacy, consent, and safeguards against unintended use.

\section*{Acknowledgements}
This work was supported in part by MSIT/IITP (No. RS-2022-II220680, RS-2020-II201821, RS-2019-II190421, RS-2024-00459618, RS-2024-00360227, RS-2024-00437633, RS-2024-00437102, RS-2025-25442569), MSIT/NRF (No. RS-2024-00357729), and KNPA/KIPoT (No. RS-2025-25393280).

%%%%%%%%%%%%%%%%%%%%%%%%%%%%%%
% \section*{Limitations}
% This document does not cover the content requirements for ACL or any
% other specific venue.  Check the author instructions for
% information on
% maximum page lengths, the required ``Limitations'' section,
% and so on.

% \section*{Acknowledgments}
% This is ack
%%%%%%%%%%%%%%%%%%%%%%%%%%%%%%

\bibliography{refernce}

%%%%%%%%%%%%%%%%%%%%%%%%%%%%%%
% \appendix
% \section{Example Appendix}
% \label{sec:appendix}
% This is an appendix.
%%%%%%%%%%%%%%%%%%%%%%%%%%%%%%

\end{document}